\documentclass[sigconf]{acmart}

\usepackage{multirow}
\usepackage{multicol}
\usepackage{algorithm}
\usepackage{algorithmic}

\AtBeginDocument{%
  \providecommand\BibTeX{{%
    \normalfont B\kern-0.5em{\scshape i\kern-0.25em b}\kern-0.8em\TeX}}}

\acmSubmissionID{1711}

\begin{document}

\title{Meta-Auxiliary Learning for Micro-Expression Recognition}


\author{Jingyao Wang, Yunhan Tian, Yuxuan Yang, Xiaoxin Chen, Changwen Zheng, Wenwen Qiang*}
\affiliation{%
  \institution{National Key Laboratory of Space Integrated Information System}
  \institution{Institute of Software Chinese Academy of Sciences}
  \institution{University of Chinese Academy of Sciences}
  \city{Beijing}
  \country{China}
  }

  





\begin{abstract}
    Micro-expressions (MEs) are involuntary movements revealing people’s hidden feelings, which has attracted numerous interests for its objectivity in emotion detection. However, despite its wide applications in various scenarios, micro-expression recognition (MER) remains a challenging problem in real life due to three reasons, including (i) data-level: lack of data and imbalanced classes, (ii) feature-level: subtle, rapid changing, and complex features of MEs, and (iii) decision-making-level: impact of individual differences. To address these issues, we propose a dual-branch meta-auxiliary learning method, called LightmanNet, for fast and robust micro-expression recognition. Specifically, LightmanNet learns general MER knowledge from limited data through a dual-branch bi-level optimization process: (i) In the first level, it obtains task-specific MER knowledge by learning in two branches, where the first branch is for learning MER features via primary MER tasks, while the other branch is for guiding the model obtain discriminative features via auxiliary tasks, i.e., image alignment between micro-expressions and macro-expressions since their resemblance in both spatial and temporal behavioral patterns. The two branches of learning jointly constrain the model of learning meaningful task-specific MER knowledge while avoiding learning noise or superficial connections between MEs and emotions that may damage its generalization ability. (ii) In the second level, LightmanNet further refines the learned task-specific knowledge, improving model generalization and efficiency. Extensive experiments on various benchmark datasets demonstrate the superior robustness and efficiency of LightmanNet.
\end{abstract}





\maketitle

\section{Introduction}
\label{sec:1}

Micro-expressions (MEs) are involuntary facial movements that are highly associated with human mental states, attitudes, and intentions \cite{merghani2018review}. These fleeting expressions offer glimpses into people's genuine feelings even in high-stakes situations, making them one of the most powerful and objective tools for understanding human psychology. The practical importance of micro-expression recognition (MER) has been demonstrated in various fields, e.g., healthcare \cite{kumar2022sentiment}, criminal interrogation \cite{nam2023facialcuenet}, and intelligent marketing \cite{zhang2023micro}.

Early methods for MER relied heavily on conventional features  \cite{gan2019off,happy2017fuzzy,huang2015facial}. Recently, with the success of deep learning in various scenarios, neural networks have received increasing interest in MER \cite{liong2019shallow,lei2020novel,mm2}. However, despite its wide applications in some fields, e.g., criminal investigations and clinical interactions involving autistic patients \cite{lu2018motion,datz2019interpretation}, automatic micro-expression recognition remains a challenging problem in real-world applications for three reasons: (i) data-level: lack of data and imbalanced classes, (ii) feature-level: subtle, rapid changing, and complex features, and (iii) decision-making-level: impact of individual differences.

\begin{figure}
    \centering
    \includegraphics[width=\linewidth]{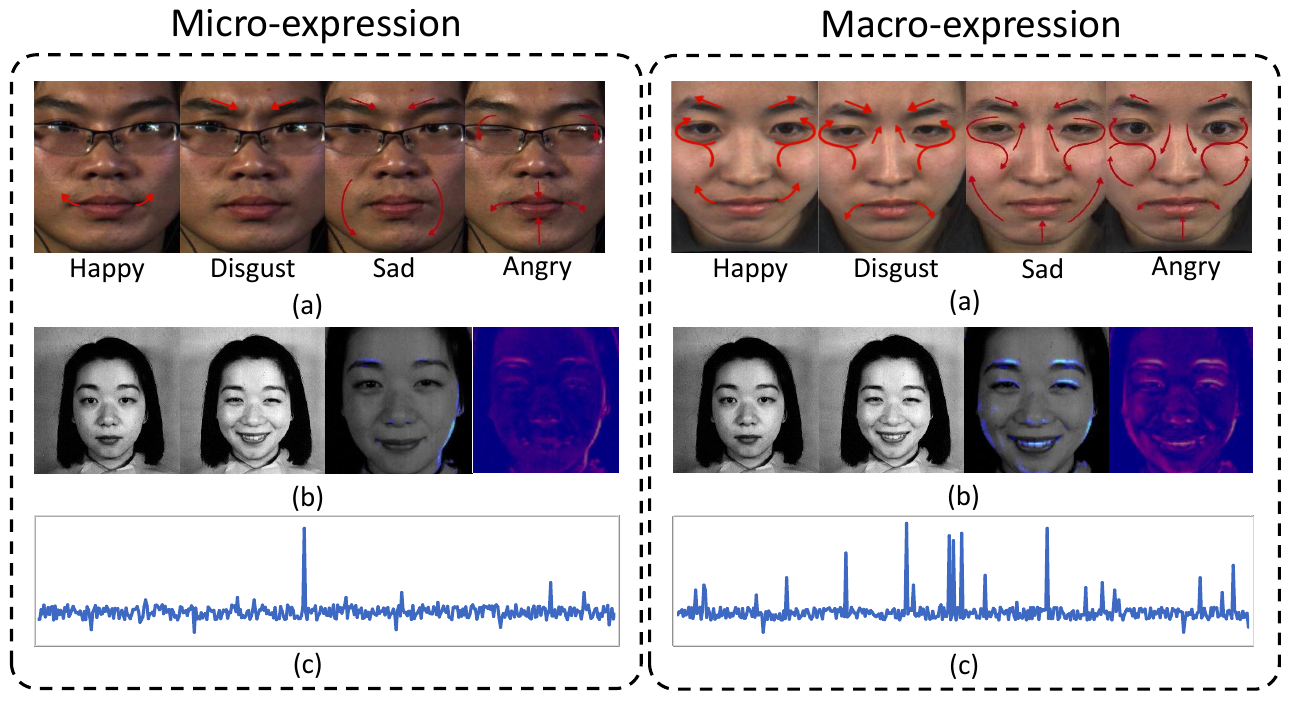}
    \caption{Comparison of micro- and macro- expressions. (a) shows examples of different expressions, where the red arrow represents the muscle movement direction. (b) shows the optical flow of expressions in a ``happy'' video sequence, where the highlight indicates the movement area. (c) shows the change in amplitude of the receptive fields of different expressions, where the spikes indicate that the pixel changes are high. More details are provided in the Appendix.}
    \label{fig:intro}
\end{figure}

Specifically, MEs are spontaneous and unplanned facial expressions revealing people’s hidden feelings in high-stakes situations when people try to conceal their true feelings, making collecting and annotating data challenging. Existing publicly available datasets are often small in size \cite{zhou2021survey,mm3}, and struggle to meet the supervised paradigm's reliance on large amounts of continuously annotated data. Next, we turn to the features and find that different from macro expressions, MEs are spontaneous, subtle, and rapid (1/25 to 1/3 second) facial movements reacting to emotional stimulus \cite{ekman2009telling,mm1}, making learning more challenging. 
Figure \ref{fig:intro} shows examples of micro- and macro-expressions, as well as facial optical flow motion and response signal under a set of emotional clips. We also provide the results including optical flow maps and dynamically changing regions. The frame where facial movement begins is denoted as the start frame, while the end frame is denoted as the offset frame, and the frame with the greatest intensity is the vertex frame. More details are provided in Appendix C. From the results, we can observe that micro-expressions have short duration and low action intensity, making their learning more challenging. In addition, MEs can also be impacted by emotional context and cultural background \cite{merghani2018review,mm4}, with individual differences. For example, some people express happiness more by raising the corners of their mouths, while others express happiness by raising their eyebrows. Therefore, MER still faces great challenges in practical applications.

To address the above issues, an ideal MER model needs to accomplish three key objectives: (i) extracting discriminative features of MER from limited data; (ii) capturing rapidly changing and imperceptible features, including the amplitude and direction of expressions, while preventing extracting noise or redundant features; (iii) acquiring general MER-related knowledge to accommodate individual differences in emotional expression.

Based on this insight, in this paper, we propose a dual-branch meta-auxiliary learning method, called LightmanNet, for robust and fast micro-expression recognition. Specifically, LightmanNet first reconstructs the MER tasks through random sampling from limited data, wherein each task encompasses distinct knowledge with both micro-expressions and macro-expressions. Subsequently, it divides these tasks into two branches: the first branch for model decision-making, handling the primary MER task, while the second branch aids in training guidance, dealing with auxiliary tasks of image amplification and alignment. These auxiliary tasks consider the similar spatial and temporal behavioral patterns between micro-expressions and macro-expressions. It aids the model in extracting more meaningful representations, steering clear of learning superficial connections that might compromise its ability to generalize. Finally, LightmanNet employs bi-level optimization to acquire general knowledge from learning in both branches, facilitating accurate MER. In summary, LightmanNet addresses the above three issues through a dual-branch bi-level optimization process, and its outstanding performance is demonstrated through extensive experiments. The contributions are as follows: 
\begin{itemize}
    \item We are the first to explore all three key challenges of micro-expression recognition in real-life multimedia applications, including data level, feature level, and decision-making level.
    \item We propose a novel dual-branch meta-auxiliary learning method, called LightmanNet, to achieve robust and fast micro-expression recognition. It enables the model to quickly obtain discriminative and general MER-related knowledge from limited data, improving model generalization and efficiency.
    \item Extensive experiments conducted on multiple benchmark datasets demonstrate the outstanding robustness and effectiveness of the proposed LightmanNet.
\end{itemize}

\section{Related Work}
\label{sec:2}

\subsection{Micro-expression Recognition}
\label{sec:2.1}

Micro-expression recognition is an important technology for facial expression analysis and sentiment analysis in real-life multimedia applications. This technology focuses on capturing short and subtle facial expression changes and discovering hidden emotional information and psychological states. It has been widely used in various fields, such as medical diagnosis \cite{kumar2022sentiment}, interpersonal communication \cite{zhang2023micro}, false information detection \cite{nam2023facialcuenet}, etc., providing richer information and deeper insights \cite{nam2023facialcuenet,kumar2022sentiment,zhang2023micro}. Existing MER methods mainly can be divided into two categories, i.e., traditional MER methods and deep learning-based MER methods.

Traditional MER methods, e.g., optical flow and local binary pattern, usually analyze micro-expression features based on features, e.g., color, texture, etc. \cite{gan2019off,happy2017fuzzy,huang2015facial}. Liu et al. \cite{Liu_Zhang_Yan_Wang_Zhao_Fu_2016} proposed a directional mean optical-flow feature with a Support Vector Machine classifier for MER. Liong et al. proposed bi-weighted oriented optical flow as a new feature extractor \cite{liong2018less}. Zhao et al. \cite{Zhao_Pietikainen_2007} applied local binary pattern-three orthogonal planes and a block-based method to analyze dynamic texture features of input images.

Deep learning-based MER methods mainly use deep neural networks for feature extraction and emotion recognition \cite{liong2019shallow,lei2020novel}. Zhai et al. \cite{zhai2023feature} proposed a self-supervised generation module to calculate the displacement of frames, utilizing a Transformer Fusion mechanism to extract features from onset-apex pairs. Nguyen et al. \cite{nebel:jair-2000} focused on interest regions of MER by detecting the differences between micro-expressions of two frames. Xia et al.  \cite{xia2021micro} propose a macro-to-micro transformation framework for micro-expression recognition, which pre-trains the model both with micro-expression data and macro-expression data respectively. 

However, although existing methods have achieved good results, they rely on a large amount of continuous and pure data while ignoring three key challenges in the real-world MER applications as mentioned in Section \ref{sec:1}. In this study, we aim to explore and solve all three key challenges of real-life MER.

\subsection{Meta-Learning}
\label{sec:2.2}

Meta-learning, also known as ``learning to learn'', attempts to adapt to new environments rapidly with limited data. The meta-learning methods can be divided into three types, including optimization-based, model-based, and metric-learning-based methods.

Optimization-based methods aim to acquire a set of optimal initialization parameters that facilitate rapid convergence when confronted with new tasks. Key methodologies in this category encompass MAML \cite{maml}, Reptile \cite{reptile}, and ANIL \cite{anil}. MAML trains a model capable of adapting to diverse tasks by sharing initial parameters across different tasks and executing multiple gradient updates. Similarly, Reptile shares initial parameters but employs an approximate update strategy, wherein the model undergoes fine-tuning through multiple iterations to approach optimal parameters. 

Model-based methods perform the optimization process in a single model. The classic methods include recurrent networks \cite{model-based-1}, convolutional networks \cite{model-based-2}, memory-augmented neural networks \cite{santoro2016meta}, and hypernetworks \cite{model-based-3, model-based-4} that embed training instances and labels of training tasks to define a predictor for test samples. In contrast to optimization-based methods, these methods offer simpler optimization procedures that do not necessitate second-order gradients. Nonetheless, there is a noted tendency for them to exhibit less robust generalization to tasks outside their trained distribution. Although they excel in few-shot learning scenarios, criticisms have arisen regarding their asymptotic performance \cite{hospedales2021meta}.

Metric-learning-based methods focus on acquiring embedding functions that map instances from distinct tasks. These functions facilitate easy classification through non-parametric methods. This concept has been investigated by various approaches, which vary in their approaches to learning embedding functions and distance metrics within the feature space. The influential methods include ProtoNet \cite{protonet}, RelationNet \cite{relationnet}, and MatchingNet \cite{vinyals2016matching}. They aim to solve problems by learning a shared embedding space and amortizing inference, where classification can be performed by calculating distances to prototype representations.

Recently, there have been studies using meta-learning to solve the problem of data scarcity in MER. Wan et al. \cite{wan2022micro} directly use MAML for micro-expression recognition. Gong et al. \cite{gong2023meta} proposed a meta-learning-based multi-model fusion network (Meta-MMFNet) to solve the existing challenging problem of the lack of training data. However, they only use meta-learning on few-shot MER scenarios under ideal conditions (i.e., relying on pure data), which may contain redundant features and noise. Despite some research \cite{dai2021cross, chen2020new, nguyen2023micron} also focused on few-shot MER, aiming to solve data-level problems, they also ignored the feature-level and decision-making-level challenges that exist in real applications as mentioned above. In this paper, We aim to propose a general framework to solve three key challenges of real-life MER, making the model obtain discriminative and generalizable MER features.

\section{Preliminary}
\label{sec:3}
Meta-learning, also known as learning to learn, seeks to bootstrap learning capabilities from a pool of tasks in order to accelerate the learning process when faced with a new task. These tasks are assumed to originate from a predetermined distribution $p(\tau)$, represented as $\tau \sim p(\tau)$. During the meta-training phase, $N$ tasks denoted as $\mathcal{D}_{tr}=\left \{ \tau_i \right \} _{i=1}^N$ are randomly selected from this distribution, and datasets associated with these tasks become available to the learner. Each task $\tau_{i}$ consists of a support set $\mathcal{D}_i^s=(X_i^s,Y_i^s)=\{ (x_{i,j}^s,y_{i,j}^s)  \}_{j=1}^{N_i^s}$ and a query set $\mathcal{D}_i^q=(X_i^q,Y_i^q)=  \{ (x_{i,j}^q,y_{i,j}^q)  \}_{j=1}^{N_i^q}$. Note that each sample $x_{i,j}^{\cdot}$ is a video sequence containing $M$ frames. In this paper, we propose introducing auxiliary tasks, i.e., image alignment between micro-expressions and macro-expressions, to guide model learning MER-related knowledge. Thus, given that our study encompasses two categories of tasks, we encounter $N$ additional auxiliary tasks $\tau^{aux} \sim p(\mathcal{T})$, each accompanied by multiple smaller datasets $\left \{ \mathcal{D}^{aux}_i \right \}_{i=1}^N $. Meta-learning strives to construct a model $f_{\theta}$ using the collective knowledge from the $2N$ training tasks ($N$ primary tasks and $N$ auxiliary tasks). The model $f_{\theta}$ is designed in a manner that enables rapid adaptation and minimization of training loss when presented with a given task.

For the learning paradigm, take the most commonly used MAML \cite{maml} as an example, it accomplishes this by acquiring initial parameters $\theta$, which involves executing a few iterations of gradient descent starting from $\theta$ utilizing the training tasks. Take the vanilla framework as an example, the model $f_{\theta}$ is optimized only based on the $N$ training tasks of the main branch without the guidance of auxiliary tasks. To obtain this starting point, MAML addresses the optimization problem: 
\begin{equation}\label{objective}
\begin{array}{l}
     \arg \min_{f_{\theta }}\frac{1}{N}\sum_{i=1}^{N}\mathcal{L}(\mathcal{D}^q _i,f_{\theta }^i), \\[8pt]
     s.t.\quad f_{\theta }^i=\arg \min_{f_\theta }\mathcal{L}(\mathcal{D}^s_i,f_{\theta }),
\end{array}
\end{equation}
where $\alpha$ is the learning rate, $f_\theta^i$ is the task-specific model obtained by taking the derivative of the meta-learning model $f_\theta$, $\mathcal{L}(\mathcal{D}^s_i,f_{\theta })$ is the training loss of the task-specific model $f_{\theta}^i$ on the support set, $\mathcal{L}(\mathcal{D}^s_i,f_{\theta })$ is the loss of the trained $f_{\theta}^i$ on the query set. Thus, the optimization process of the meta-learning $f_\theta$ can be viewed as calculating the second derivative of $f_\theta$.

\section{Method}
\label{sec:4}
\begin{figure*}[t]
    \centering
    \includegraphics[width=\linewidth]{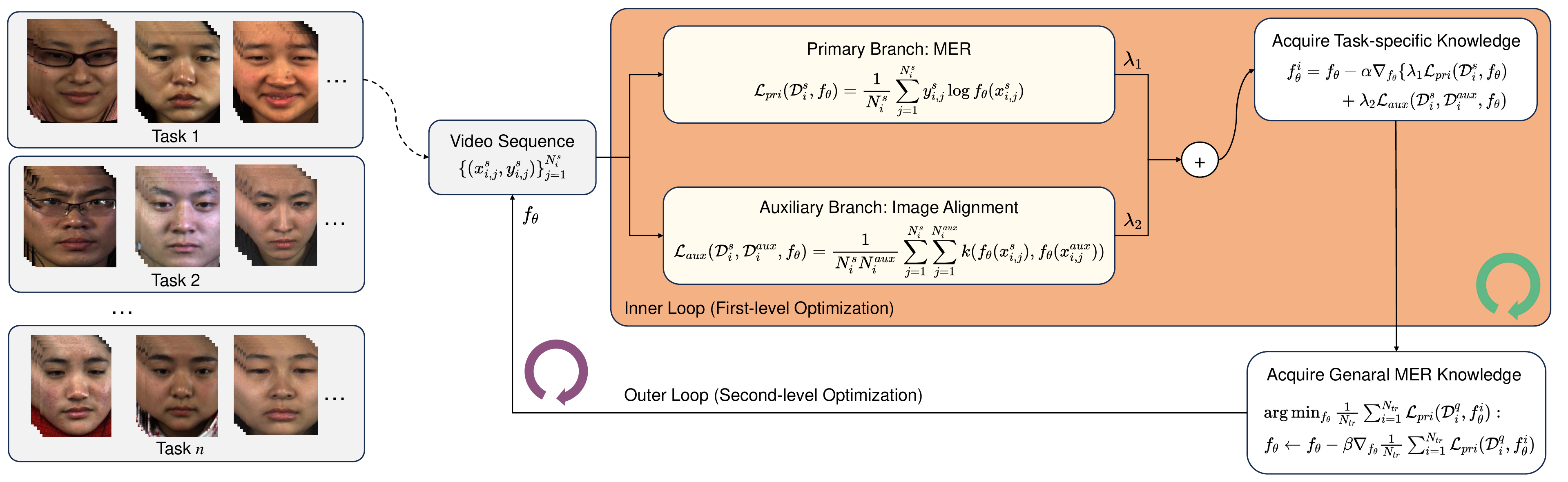}
    \caption{Overview of LightmanNet. It first builds various training tasks (gray boxes on the left), and then learns general MER-related knowledge via bi-level optimization: (i) in the first level (green circle), the model $f_{\theta}$ learns task-specific MER knowledge for task $\tau_i$ and obtain $f_{\theta}^i$ through two branches of learning, i.e., $f_{\theta}\to f_{\theta}^i$, where the primary branch is used to learn MER features, while the auxiliary branch is used to guide $f_{\theta}$ to obtain discriminative features; then, (ii) in the second level (purple circle), $f_{\theta}$ refines the learned task-specific knowledge of $f_{\theta}^i$, i.e., optimized with the cumulative loss of multiple $f_{\theta}^i$.}
    \label{fig:framework}
\end{figure*}

In this section, we present the proposed method, LightmanNet, for micro-expression recognition. We first introduce the model architecture of LightmanNet in Subsection \ref{sec:4.1}, which has a two-branch architecture that jointly solves two related tasks. Given a sequence of observed frames, the primary task aims to recognize the emotions contained in the micro-expressions of these frames. In addition to the primary branch, LightmanNet has another branch that solves an auxiliary task complementary to the primary task, i.e., image alignment. These two tasks share the backbone features and can be jointly learned. Next, we introduce the learning details of LightmanNet in Subsection \ref{sec:4.2}, which is a dual-branch bi-level optimization process of meta-auxiliary learning. Figure \ref{fig:framework} provides an overview of our approach. The pseudo-code is shown in Algorithm \ref{alg:algorithm}.

\begin{figure}[t]
    \centering
    \includegraphics[width=\linewidth]{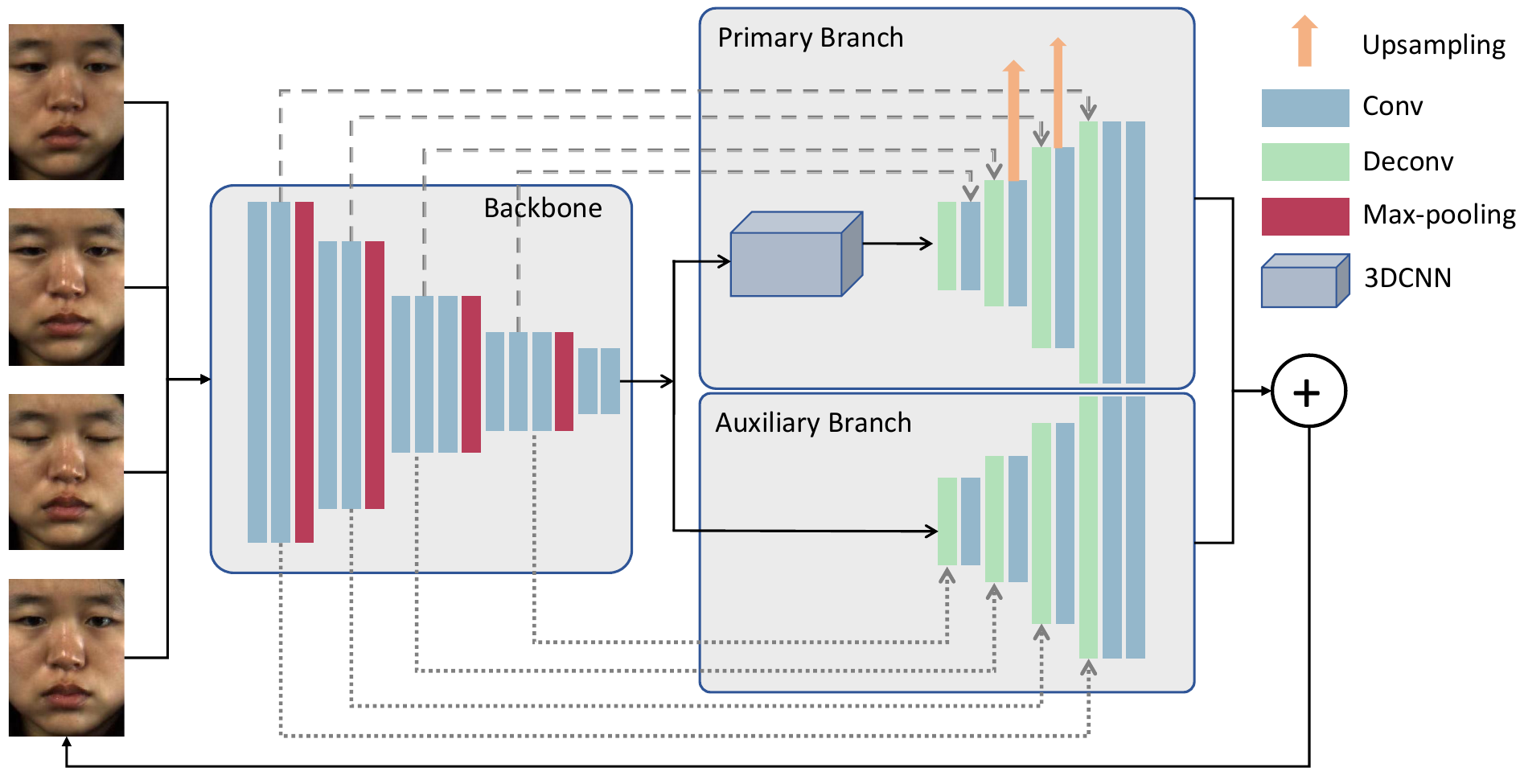}
    \caption{Illustration of our model architecture. The two branch structures of the model share the same 2DCNN network for feature extraction (left side), but the last layer of the encoder and the decoder are different (right side).}
    \label{fig:structure}
\end{figure}

\subsection{Model Architecture}
\label{sec:4.1}
As shown in Figure \ref{fig:structure}, LightmanNet has a dual-branch structure. The network takes multiple consecutive video frames as input, and we denote the $M$ frames of each input video sequence as $\left \{ I_i \right \}_{i=1}^M$. The main branch extracts the emotions from micro-expressions in the video sequences. The auxiliary branch contains image alignment tasks between micro-expressions and macro-expressions for their resemblance in both spatial and temporal behavioral patterns. 

These two branches share the same 2DCNN backbone network for feature extraction, which can be any image classification network, e.g., Conv4 \cite{maml}, VGG-19 \cite{vgg}, ResNet \cite{resnet}, and DenseNet \cite{densenet}. To save computation, we use the simple and lightweight Conv4 with the last fully convolutional layer removed as our backbone. Then, we combine a 3DCNN network \cite{3dcnn} to encode the multiple-frame features of the last convolutional layer of the backbone network, and use a single 3D convolutional layer with kernel size 6 along the time dimension. The 3DCNN module can capture the temporal information in the video sequences. Finally, we design the decoder following \cite{u-net,liu2023meta}, when it upsamples the features from low resolution to input resolution, a convolutional layer is added to output three different scales of the feature maps. We explain the two branches in more detail below.

\textbf{Primary Task Learning}. We use a commonly used classification loss, i.e., cross-entropy loss \cite{de2005tutorial}, to constrain the learning of the main task (MER tasks) branch:
\begin{equation}\label{loss_pri}
    \mathcal{L}_{pri}(\mathcal{D}^s_i,f_{\theta })= \frac{1}{N_i^s} \sum_{j=1}^{N_i^s}y_{i,j}^s\log {f_\theta }(x_{i,j}^s),
\end{equation}
where $\mathcal{L}_{pri}(\tau_i)$ is the loss for the primary MER task $\tau_i$, $\mathcal{D}^s_i$ is the support set of task $\tau_i$, $f_\theta $ is the learning model.

\textbf{Auxiliary Task Learning}. Learning a primary task alongside a proper auxiliary task can force the model to capture more meaningful representations, steering clear of learning superficial connections that might compromise its ability to generalize \cite{navon2020auxiliary,liu2023meta}. The auxiliary task should be selected wisely to support the primary task, or else the primary task's performance will deteriorate. We need an auxiliary task that can teach the network how to recognize subtle, dynamic, and complex ME features. 

In this paper, we propose to use image alignment \cite{szeliski2007image,xia2021micro} as the auxiliary task. Image alignment is performed since micro-expression and macro-expression share some similarities in both spatial and temporal behavior patterns. To perform well in this task, the model needs to learn feature representations that capture the geometric and semantic information of the scene, which intuitively will be useful for efficient MER as well.

We design the image alignment branch similar to the MER decoder based on the shared feature encoder.
In this branch, we evaluate the feature alignment degree of micro-expressions and macro-expressions by measuring Maximum Mean Discrepancy (MMD). Given the support set $\mathcal{D}_{i}^s $ of the main task $\tau_i$ and the support set$\mathcal{D}_{i}^{aux} $ of the auxiliary task $\tau_i^{aux}$, the loss can be expressed as:
\begin{equation}\label{loss_aux}
\begin{array}{l}
\scalebox{0.85}{$
     \mathcal{L}_{aux}(\mathcal{D}^s_i, \mathcal{D}^{aux}_i, f_{\theta})=\frac{1}{N_i^sN_i^{aux}} \sum_{j=1}^{N_i^s} \sum_{j=1}^{N_i^{aux}} k(f_{\theta}(x_{i,j}^s),f_{\theta}(x_{i,j}^{aux}))$}, \\[8pt]
     s.t. \quad k(x, y) = \exp\left(-\frac{\|x - y\|^2}{2\sigma^2}\right) ,
\end{array}
\end{equation}
where $N_i^s$ and $N_i^{aux}$ are the number of samples in tasks $\tau_i$ and $\tau_i^{aux}$, $\mathcal{D}^s_i$ and $\mathcal{D}^{aux}_i$ are the support sets of tasks $\tau_i$ and $\tau_i^{aux}$, and $k(\cdot, \cdot)$ is the Gaussian kernel function.

\subsection{Meta-Auxiliary Learning}
\label{sec:4.2}
To effectively extract valuable knowledge from limited data, LightmanNet utilizes a bi-level optimization for meta-auxiliary learning. This approach enables the acquisition of general knowledge from both branches, thereby enhancing accurate Micro-Expression Recognition (MER). Specifically, LightmanNet first extracts micro-expression features from the limited data of each task through joint training on two branches, facilitating the acquisition of meaningful knowledge. Subsequently, LightmanNet employs knowledge distillation on the acquired knowledge to further enhance the model's generalization capability.

Formally, The goal of LightmanNet is to learn the parameters $\theta$ of the model $f_{\theta}$ so that they can be effectively used for adaptation. The model $f_{\theta}$ is divided into three parts as shown in Figure \ref{fig:structure}, expressed as $f_{\theta }=g\circ \left \{ h_{pri},h_{aux} \right \} $, where $g$ represents the backbone network, $h_{pri}$ and $h_{aux}$ denote the primary MER branch and the auxiliary image alignment branch, respectively. In the first level optimization, $f_{\theta}$ aims to learn task-specific knowledge by minimizing the losses $\mathcal{L}_{pri}$ and $\mathcal{L}_{aux}$ in Eq.\ref{loss_pri} and Eq.\ref{loss_aux} over tasks $\tau_i$ and $\tau_i^{aux}$ of both the branches. Then, in the second level optimization, $f_{\theta}$ is updated based on the cumulative loss of multiple groups of tasks, that is, acquiring general knowledge distilled from the acquired task-specific knowledge. The objective of LightmanNet becomes:
\begin{equation}\label{eq:lightman_objective}
\begin{array}{l}
     \arg \min _{f_{\theta }}\frac{1}{N_{tr}}\sum_{i=1}^{N_{tr}}\mathcal{L}_{pri}(\mathcal{D}^q _i,f_{\theta }^i), \\[8pt]
     s.t. \quad \scalebox{0.9}{$f_{\theta }^i=\arg \min_{f_{\theta }}\lambda_1 \mathcal{L}_{pri}(\mathcal{D}^s_i,f_{\theta })+\lambda_2\mathcal{L}_{aux}(\mathcal{D}^s_i, \mathcal{D}^{aux}_i, f_{\theta})$},
\end{array}
\end{equation}
where $N_{tr}$ denotes the number of training tasks, $\lambda_1$ and $\lambda_2$ are the weights of the losses $\mathcal{L}_{pri}$ (Eq.\ref{loss_pri}) and $\mathcal{L}_{aux}$ (Eq.\ref{loss_aux}) over tasks $\tau_i$ and $\tau_i^{aux}$, $f_\theta^i$ is the task-specific model obtained by taking the derivative of the model $f_\theta$. The learning rates of the first-level and second-level objectives are $\alpha$ and $\beta$, respectively. It is worth noting that the auxiliary task is used to guide the model to extract complex micro-expression features and does not involve the acquisition of general knowledge, i.e., the second-level optimization. 

Through the above process, we obtained the final model $f_{\theta}$ which can quickly adapt to any task well with limited data. In the testing phase, for any unseen MER task $\tau_{te}$, $f_{\theta}$ is first fine-tuned on the data set (usually only 1 step is required) to quickly extract task-specific MER knowledge And get $f_{\theta}^{te}$. Finally, we process the unseen video data in the test task based on the model $f_{\theta}^{te}$ to obtain accurate MER results. Therefore, LightmanNet well eliminates the effects of lack of samples, complex features, and individual differences, and its generalization and robustness will be verified in Section \ref{sec:5}.

\begin{algorithm}[t]
	\caption{Pseudo-Code of LightmanNet}
	\label{alg:algorithm}
	\begin{algorithmic}
		\REQUIRE The primary tasks $\mathcal{D}_{tr}$; the auxiliary tasks $\mathcal{D}^{aux}$; meta-auxiliary learning model $f_{\theta}$; The hyperparameter $\lambda_1$ and $\lambda_2$; Learning rates $\alpha$ and $\beta$\;
            \FOR{all $\tau_i$}
            \STATE Obtain support sets and query sets for both the primary MER task and the auxiliary task
            \STATE Calculate primary task loss $\mathcal{L}_{pri}$ via Eq.2
            \STATE Calculate auxiliary task loss $\mathcal{L}_{aux}$ via Eq.3
            \STATE Update the task-specific model $f^i_{\theta}$ via 4
            \ENDFOR
            \STATE Update the model $f_{\theta}$ via Eq.4
            \STATE \textbf{return} solution
	\end{algorithmic} 
\end{algorithm}

\begin{table*}[t]
  \centering
    \caption{Performance (accuracy \& F1 score $\pm$ 95\% confidence interval) of baselines and LightmanNet on the datasets mentioned in Subsection \ref{sec:5.1}. All results are provided in the corresponding papers or are obtained based on the source code. Note that the meta-learning-based baselines follow the few-shot settings mentioned in Subsection \ref{sec:5.3}. The ``-'' denotes that the result is not reported, and the best results are highlighted in \textbf{bold}. See Appendix C for full results.}
\label{tab:comparison}
  \resizebox{\textwidth}{!}{
  \begin{tabular}{l|cc|cc|cc|cc|cc}
    \toprule
   \multirow{2}{*}{\textbf{Method}} & \multicolumn{2}{c|}{\textbf{SAMM(CK+)}} & \multicolumn{2}{c|}{\textbf{SMIC(CK+)}} & \multicolumn{2}{c|}{\textbf{CASME(CK+)}} & \multicolumn{2}{c|}{\textbf{CASME II(CK+)}} & \multicolumn{2}{c}{\textbf{CAS(ME)²(CK+)}}\\
    & \textbf{ACC} & \textbf{F1} & \textbf{ACC} & \textbf{F1} & \textbf{ACC} & \textbf{F1} & \textbf{ACC} & \textbf{F1} & \textbf{ACC} & \textbf{F1} \\
    \midrule
    LBP-TOP & 59.23 $\pm$ 0.35& 36.40 $\pm$ 0.42& 58.80 $\pm$ 0.35& 60.05 $\pm$ 0.41 & 57.11 $\pm$ 0.49 & 58.02 $\pm$ 0.45 & 49.00 $\pm$ 0.53& 51.01 $\pm$ 0.34 & - & - \\
    LBP-SIP  & 41.56 $\pm$ 0.51& 40.60 $\pm$ 0.51& 44.53 $\pm$ 0.52& 44.92 $\pm$ 0.34& 43.42 $\pm$ 0.36& 45.30 $\pm$ 0.34& 46.53 $\pm$ 0.47& 44.82 $\pm$ 0.48& - & - \\
    STLBP-IP& 56.82 $\pm$ 0.38& 52.73 $\pm$ 0.40& 57.96 $\pm$ 0.35& 58.00 $\pm$ 0.36& 56.72 $\pm$ 0.29& 57.92 $\pm$ 0.53& 59.53 $\pm$ 0.35& 57.02 $\pm$ 0.32& 51.73 $\pm$ 0.42& 52.33 $\pm$ 0.38\\
    STCLQP& 63.82 $\pm$ 0.32& 61.11 $\pm$ 0.41& 58.31 $\pm$ 0.35& 58.38 $\pm$ 0.51& 57.31 $\pm$ 0.32& 56.02 $\pm$ 0.27& 64.10 $\pm$ 0.33& 63.88 $\pm$ 0.37& 56.31 $\pm$ 0.33& 56.87 $\pm$ 0.32\\
    HIGO& - & - & 68.20 $\pm$ 0.33& 67.5 $\pm$ 0.33& 61.52 $\pm$ 0.33& 63.20 $\pm$ 0.34& - & - & 58.26 $\pm$ 0.34& 58.10 $\pm$ 0.28\\
    FHOFO& - & - & 51.89 $\pm$ 0.40& 52.34 $\pm$ 0.42& 67.01 $\pm$ 0.29& 54.92 $\pm$ 0.34& 56.60 $\pm$ 0.34& 52.49 $\pm$ 0.33& 48.93 $\pm$ 0.25& 47.92 $\pm$ 0.30\\
    Bi-WOOF& 58.30 $\pm$ 0.36& 39.72 $\pm$ 0.49& 62.22 $\pm$ 0.33& 62.00 $\pm$ 0.29& 69.42 $\pm$ 0.38& 69.8 $\pm$ 0.23& 58.87 $\pm$ 0.36& 61.10 $\pm$ 0.36& 50.62 $\pm$ 0.33& 51.30 $\pm$ 0.34\\
    \midrule 
    OFF-Apex& 68.11 $\pm$ 0.32& 54.22 $\pm$ 0.36& 67.66 $\pm$ 0.30& 67.03 $\pm$ 0.27& 79.22 $\pm$ 0.25& 78.5 $\pm$ 0.23& - & - &  - &  - \\
    Boost&  - &  - & 68.89 $\pm$ 0.31& 68.1 $\pm$ 0.39& - & - & 70.92 $\pm$ 0.30& 71.90 $\pm$ 0.30& 62.84 $\pm$ 0.35& 63.23 $\pm$ 0.31\\
    DSSN& 57.35 $\pm$ 0.35& 46.43 $\pm$ 0.51& 63.43 $\pm$ 0.32& 64.68 $\pm$ 0.48& - & - & 70.80 $\pm$ 0.30& 73.10 $\pm$ 0.31&  - & - \\
    AU-GACN& 70.24 $\pm$ 0.30& 43.33 $\pm$ 0.34& - & - & 80.26 $\pm$ 0.33& 80.10 $\pm$ 0.21& 71.22 $\pm$ 0.29& 35.54 $\pm$ 0.39& - &  - \\
    Dynamic& - & - & 76.10 $\pm$ 0.28& 71.01 $\pm$ 0.35& 81.80  $\pm$ 0.26& 77.00 $\pm$ 0.30& 72.61 $\pm$ 0.28& 67.09 $\pm$ 0.30& 65.18 $\pm$ 0.32& 65.15 $\pm$ 0.36\\
 MicroNet& 74.11 $\pm$ 0.28& 73.64 $\pm$ 0.48& 76.80 $\pm$ 0.28& 74.45 $\pm$ 0.45& 80.59 $\pm$ 0.32& 81.79 $\pm$ 0.32& 75.61 $\pm$ 0.29& 70.12 $\pm$ 0.30&  - & - \\
 Graph-TCN& 75.00 $\pm$ 0.28& 69.93 $\pm$ 0.37& 77.46 $\pm$ 0.27& 76.19 $\pm$ 0.45& 79.61 $\pm$ 0.39& 84.11 $\pm$ 0.40& 74.08 $\pm$ 0.29& 72.58 $\pm$ 0.32& 67.43 $\pm$ 0.26&68.73 $\pm$ 0.39\\
 FR& 60.13 $\pm$ 0.35& 61.56 $\pm$ 0.35& 57.90 $\pm$ 0.35& 57.33 $\pm$ 0.35& 77.54 $\pm$ 0.27& 77.80 $\pm$ 0.33& 68.38 $\pm$ 0.31& 68.78 $\pm$ 0.37& 61.58 $\pm$ 0.33&61.35 $\pm$ 0.40\\
 SelfME & 69.43 $\pm$ 0.28 & 70.22 $\pm$ 0.31 & - & - & 75.25 $\pm$ 0.30 & 74.66 $\pm$ 0.19 & 74.55 $\pm$ 0.29 & 76.30 $\pm$ 0.29 & 71.01 $\pm$ 0.22 & 69.77 $\pm$ 0.25 \\
 FRL-DGT & - & - & 73.08 $\pm$ 0.32 & 72.01 $\pm$ 0.32 & 78.00 $\pm$ 0.26 & 79.13 $\pm$ 0.22 & 75.12 $\pm$ 0.30 & 76.58 $\pm$ 0.22 & 70.11 $\pm$ 0.25 & 70.56 $\pm$ 0.25 \\
 MiMaNet& 76.70 $\pm$ 0.26& 76.42 $\pm$ 0.28& 78.60 $\pm$ 0.28& 77.82 $\pm$ 0.28& 81.21 $\pm$ 0.33& 83.22 $\pm$ 0.32& 79.90 $\pm$ 0.28& 75.93 $\pm$ 0.34&  - & - \\
 AMAN& 68.85 $\pm$ 0.31& 66.83 $\pm$ 0.31& 79.87 $\pm$ 0.27& 77.14 $\pm$ 0.31& - & - & 75.40 $\pm$ 0.28& 71.33 $\pm$ 0.34& 67.45 $\pm$ 0.32&66.49 $\pm$ 0.43\\
 $\mu$-BERT & 80.16 $\pm$ 0.25 & 78.41 $\pm$ 0.22 & 81.38 $\pm$ 0.29 & 80.56 $\pm$ 0.31 & 83.84 $\pm$ 0.33 & 84.09 $\pm$ 0.30 & 85.00 $\pm$ 0.29 & 85.36 $\pm$ 0.22 & 80.01 $\pm$ 0.28 & 81.33 $\pm$ 0.25 \\
 \midrule 
 Meta-MMFNet& 64.15 $\pm$ 0.24 & 66.51 $\pm$ 0.37 & 63.12 $\pm$ 0.33& 64.93 $\pm$ 0.23& 69.51 $\pm$ 0.30& 67.67 $\pm$ 0.33& 80.92 $\pm$ 0.27& 81.23 $\pm$ 0.30&  72.14 $\pm$ 0.44 & 73.31 $\pm$ 0.55 \\
 MEReMAML & 74.23 $\pm$ 0.30 & 76.37 $\pm$ 0.31 & 80.78 $\pm$ 0.24 & 81.51 $\pm$ 0.32 & 78.12 $\pm$ 0.30 & 79.56 $\pm$ 0.28 & 86.21 $\pm$ 0.20 & 89.26 $\pm$ 0.21 & 76.12 $\pm$ 0.29 & 75.89 $\pm$ 0.18 \\
 \midrule
 \textbf{LightmanNet}& \textbf{81.83 $\pm$ 0.44} & \textbf{82.93 $\pm$ 0.35}& \textbf{85.19 $\pm$ 0.34}& \textbf{83.83 $\pm$ 0.34}& \textbf{87.83 $\pm$ 0.60}& \textbf{89.78 $\pm$ 0.59}& \textbf{93.48 $\pm$ 0.42} & \textbf{90.49 $\pm$ 0.47}& \textbf{85.23 $\pm$ 0.26} & \textbf{82.22 $\pm$ 0.29}\\
  \bottomrule
  \end{tabular}}
\end{table*}

\begin{table*}[t]
  \centering
  \caption{Performance comparison (accuracy \& F1 score $\pm$ 95\% confidence interval) of few-shot MER. The ``*'' denotes that the results are recorded on a subset instead of recorded via five rounds. See Appendix C for full results.}
 \label{tab:few-shot}
  \resizebox{\textwidth}{!}{
  \begin{tabular}{l|cc|cc|cc|cc|cc}
    \toprule
    \multirow{2}{*}{\textbf{Method}} & \multicolumn{2}{c|}{\textbf{SAMM(CK+)}} & \multicolumn{2}{c|}{\textbf{SMIC(CK+)}} & \multicolumn{2}{c|}{\textbf{CASME(CK+)}} & \multicolumn{2}{c|}{\textbf{CASME II(CK+)}} & \multicolumn{2}{c}{\textbf{CAS(ME)²(CK+)}}\\
    & \textbf{ACC} & \textbf{F1} & \textbf{ACC} & \textbf{F1} & \textbf{ACC} & \textbf{F1} & \textbf{ACC} & \textbf{F1} & \textbf{ACC} & \textbf{F1} \\
    \midrule
    MAML & 59.00 $\pm$ 0.00 & 61.00 $\pm$ 0.00 & 58.00 $\pm$ 0.00 & 57.00 $\pm$ 0.00 & 64.00 $\pm$ 0.00 & 61.00 $\pm$ 0.00 & 63.00 $\pm$ 0.00 & 60.00 $\pm$ 0.00 & 59.00 $\pm$ 0.00 & 62.00 $\pm$ 0.00 \\
    Reptile &  58.32 $\pm$ 0.28&  59.75 $\pm$ 0.34&  57.27 $\pm$ 0.43&  56.22 $\pm$ 0.38&  62.76 $\pm$ 0.38&  62.61 $\pm$ 0.39&  62.42 $\pm$ 0.19&  62.34 $\pm$ 0.78&  60.96 $\pm$ 0.48&  63.11 $\pm$ 0.32\\
    ANIL &  57.47 $\pm$ 0.37&  62.42 $\pm$ 0.39&  -&  -&  64.53 $\pm$ 0.33&  61.97 $\pm$ 0.12&  -&  -&  58.84 $\pm$ 0.32&  62.76 $\pm$ 0.34\\
    ProtoNet &  59.97 $\pm$ 0.33&  64.53 $\pm$ 0.26&  60.22 $\pm$ 0.31&  57.62 $\pm$ 0.25&  63.35 $\pm$ 0.43&  64.28 $\pm$ 0.43&  63.27 $\pm$ 0.76&  59.32 $\pm$ 0.46&  59.87 $\pm$ 0.43&  61.65 $\pm$ 0.31\\
    RelationNet &  -&  -&  58.11 $\pm$ 0.27&  56.37 $\pm$ 0.21&  -&  -&  64.63 $\pm$ 0.54&  61.25 $\pm$ 0.32&  60.76 $\pm$ 0.21&  61.96 $\pm$ 0.38\\
    MatchingNet &  61.42 $\pm$ 0.36&  59.51 $\pm$ 0.32&  60.46 $\pm$ 0.34&  58.79 $\pm$ 0.38&  65.43 $\pm$ 0.76&  61.74 $\pm$ 0.35&  62.54 $\pm$ 0.35&  61.45 $\pm$ 0.76&  -&  -\\
    \midrule
    Meta-MMFNet& 64.15 $\pm$ 0.24 & 66.51 $\pm$ 0.37 & 63.12 $\pm$ 0.33& 64.93 $\pm$ 0.23& 69.5 $\pm$ 0.30& 67.67 $\pm$ 0.33& 80.92 $\pm$ 0.27& 81.23 $\pm$ 0.30&  72.14 $\pm$ 0.44 & 73.31 $\pm$ 0.55 \\
    MEReMAML& 79.03 $\pm$ 0.41 & 77.12 $\pm$ 0.39 & 79.36 $\pm$ 0.06$^*$ & 81.23 $\pm$ 0.19$^*$ & 81.22 $\pm$ 0.49 & 79.15 $\pm$ 0.33 & 92.97 $\pm$ 0.54& 91.05 $\pm$ 0.44 & 84.16 $\pm$ 0.56 & 80.19 $\pm$ 0.32 \\
    \midrule
    \textbf{LightmanNet} & \textbf{81.83 $\pm$ 0.44} & \textbf{82.93 $\pm$ 0.35} & \textbf{85.19 $\pm$ 0.34}& \textbf{83.83 $\pm$ 0.34}& \textbf{87.83 $\pm$ 0.60}& \textbf{89.78 $\pm$ 0.59}& \textbf{93.48 $\pm$ 0.42} & \textbf{90.49 $\pm$ 0.47}& \textbf{85.23 $\pm$ 0.26}&\textbf{82.22 $\pm$ 0.29}\\
  \bottomrule
  \end{tabular}
  }
\end{table*}

\section{Experiments}
\label{sec:5}
In this section, we first introduce the experimental settings in Subsection \ref{sec:5.1}. Then, we present the comparative experiments between our method and the baselines from two aspects: performance and model efficiency in Subsection \ref{sec:5.2}. Next, we analyze the robustness of our approach in Subsection \ref{sec:5.3}, which reflects the effect of the proposed method when lack of data and differences between classes. Furthermore, we perform ablation studies to further explore how LightmanNet works in Subsection \ref{sec:5.4}. All results are averaged over five rounds of testing. More details and additional experiments are provided in the Appendix.

\subsection{Experimental Settings}
\label{sec:5.1}
In this subsection, we introduce the datasets, baselines, implementation details, and evaluation metrics in sequence.

  \paragraph{Datasets} We select five MER benchmark datasets, including (i) SAMM dataset \cite{davison2016samm} which contains 159 micro-expression clips from 32 participants; (ii) SMIC dataset \cite{li2013spontaneous} that consists of 164 micro-expression samples from 16 subjects; (iii) CASME dataset \cite{yan2013casme} that contains 195 micro-expression videos sampled from 19 individuals; (iv) CASME II \cite{yan2014casme} that expands on the CASME and contains five emotions; and (v) CAS(ME)² dataset \cite{qu2017cas} includes both macro-expressions and micro-expressions, categorizing emotions into three groups: positive, negative, and surprise, and be further divided into 12 detailed emotional labels, i.e., happiness, sadness, fear, surprise, anger, disgust, trust, hate, contempt, guilt, grateful, and bittersweet \cite{wang2023amsa,mm5,mm6}. Meanwhile, we use the Composite Database Evaluation (CDE) which combines samples from all datasets together into a single database to evaluate the performance of our proposed LightmanNet following \cite{nguyen2023micron}. For auxiliary tasks construction, we use the CK+ database as the macro-expression database following \cite{xia2021micro}. More details are shown in Appendix A.

\paragraph{Baselines} We extensively compare LightmanNet with four types of baselines for comprehensive evaluation, including (i) traditional MER methods \cite{gong2023meta} including LBP-TOP \cite{le2016sparsity}, LBP-SIP \cite{wang2015lbp}, STLBP-IP \cite{huang2015facial}, STCLQP \cite{Huang_Zhao_Hong_Zheng_Pietikäinen_2016}, HIGO \cite{li2017towards}, FHOFO \cite{happy2017fuzzy} and Bi-WOOF \cite{liong2018less}; (ii) deep-learning-based MER methods \cite{nguyen2023micron} including OFF-Apex \cite{gan2019off}, Boost \cite{peng2019boost}, DSSN \cite{khor2019dual}, AU-GACN \cite{xie2020assisted}, Dynamic \cite{sun2020dynamic}, MicroNet \cite{xia2020learning}, Graph-TCN \cite{lei2020novel}, MiMaNet \cite{xia2021micro}, FR \cite{zhou2022feature}, AMAN \cite{wei2022novel}, SelfME \cite{fan2023selfme}, $\mu$-BERT \cite{nguyen2023micron} and FRL-DGT \cite{zhai2023feature} ; (iii) meta-learning methods mentioned in Subsection \ref{sec:2.2}, e.g., MAML \cite{maml}, Reptile \cite{reptile} and ANIL \cite{anil}; and (iv) meta-learning-based MER methods including MEReMAML \cite{wan2022micro} and Meta-MMFNet \cite{gong2023meta}. These baselines cover the SOTA baselines and classic MER methods for each benchmark dataset. More baselines are provided in Appendix B.

\paragraph{Implementation Details} We choose the Conv4 backbone for the encoder and the overall structure is illustrated in Subsection \ref{sec:4.1}. Following the convolution and filtering stages, we apply batch normalization, ReLU non-linear activation, and $2 \times 2$ max pooling in sequence. The output of this encoder consists of two separate branches, i.e., the primary branch and the auxiliary branch. The construction of training tasks of each dataset or domain follows the process mentioned in Section \ref{sec:3}. Moving to the optimization stage, we employ the Adam optimizer with momentum and weight decay values set at 0.9 and $10^{-4}$. The initial learning rates, i.e., $\alpha$ and $\beta$ mentioned in Eq.\ref{eq:lightman_objective}, for all experiments are established at 0.2 and 0.1, with the option for linear scaling as required. All experiments are conducted on 10 NVIDIA V100 GPUs.

\paragraph{Evaluation Metrics} 
We select five commonly used metrics for evaluation, including (i) accuracy (\%) which is a commonly used metric to evaluate the overall performance of a classification model; (ii) F1 score (\%) which combines precision and recall into a single value; (iii) unweighted average recall (UAR) (\%) which computes the average recall across all classes without considering the class imbalance; (iv) training time (h) which is used to evaluate the model efficiency; and (v) model size (M) refers to the amount of memory or storage space required to store the model. All results reported are the averages of five independent runs.

\begin{figure}[t]
    \centering
    \includegraphics[width=0.85\linewidth]{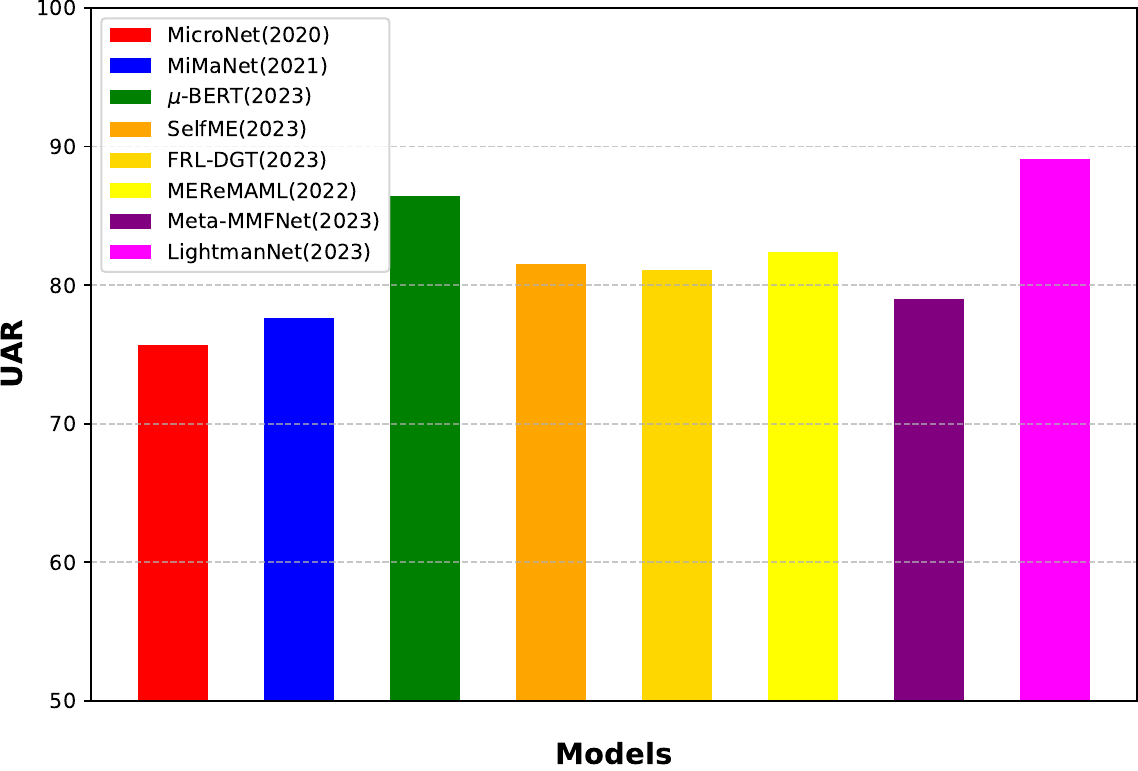}
    \caption{Performance comparison (UAR) of the baselines and our proposed LightmanNet on the CDE dataset. We choose the recently proposed baselines for comparison, which cover the SOTA and classic MER method on the CDE dataset, including meta-learning-based and deep-learning-based.}
    \label{fig:test}
\end{figure}

\begin{figure}[t]
    \centering
    \includegraphics[width=0.85\linewidth]{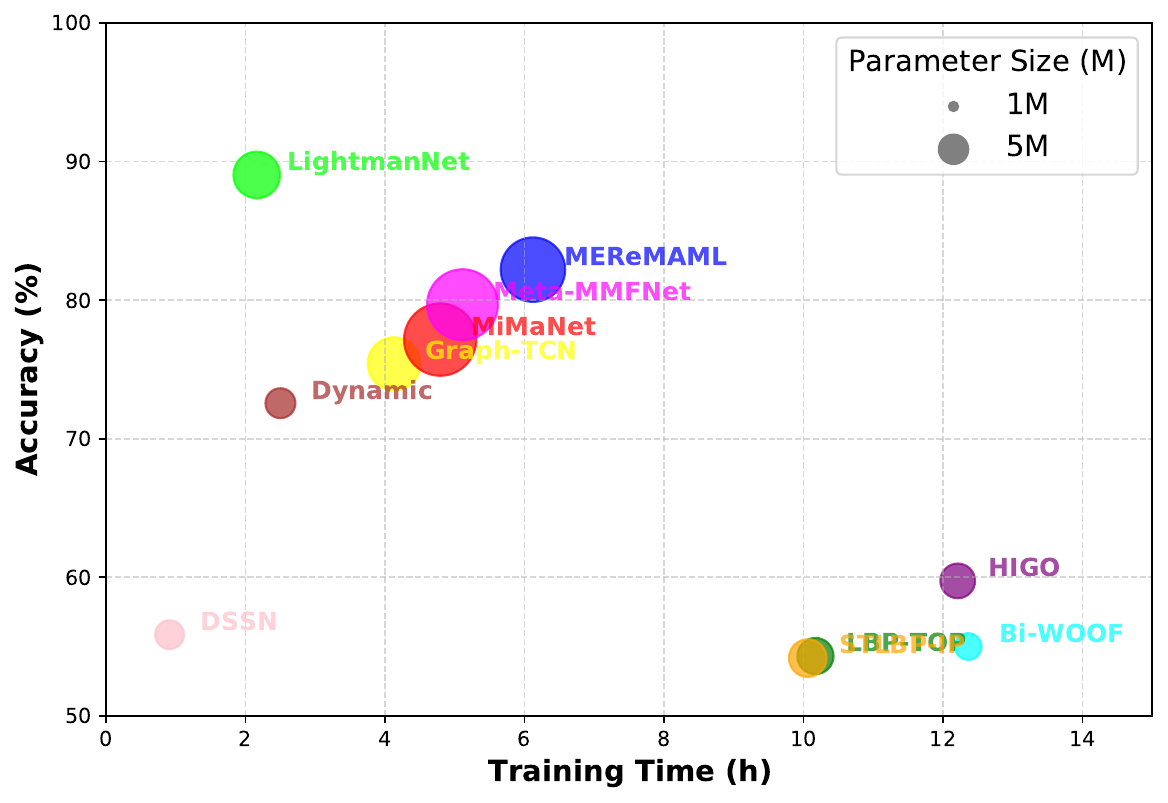}
    \caption{Model efficiency comparison of the baselines and LightmanNet on the CDE dataset, which is recorded with the same batch size and official code configuration. We record the performance of methods cover all the four types of baselines in Section \ref{sec:5.1} for 5 rounds of experiments.}
    \label{fig:model_eff}
\end{figure}

\begin{table*}
\centering
\caption{Ablation study of LightmanNet on the  CDE dataset, i.e., we analyze the impact of the three components of LightmanNet by activate or deactivate/replace the corresponding modules, including the feature extractor, the guidance of auxiliary task (the effect of introducing auxiliary tasks), and bi-level optimization. The ``$\checkmark$" indicates that the corresponding module is activated in the current experiment. We record the average results from five rounds of testing}
  \label{tab:abla_module}
  \resizebox{0.92\textwidth}{!}{
  \begin{tabular}{l|l|cccccc}
    \toprule
    \multirow{4}{*}{Feature extractors} & Conv4     &$\checkmark$&$\checkmark$&$\checkmark$&  &  &  \\
    & ResNet  &  &  &  & $\checkmark$ &  &  \\
    & VGG-19  &  &  &  &  &$\checkmark$ &  \\
    & DenseNet  &  &  &  &  &  &$\checkmark$ \\
    \cline{1-2}   
    \multicolumn{2}{l|}{The guidance of auxiliary tasks}  &$\checkmark$ & &$\checkmark$&$\checkmark$&$\checkmark$&$\checkmark$\\   
    \multicolumn{2}{l|}{Bi-level optimization}  &  & $\checkmark$ &$\checkmark$ &$\checkmark$&$\checkmark$&$\checkmark$ \\
    \midrule    
    \multicolumn{2}{l|}{ACC} & 80.15 $\pm$ 0.33 & 83.36 $\pm$ 0.38 & 91.10 $\pm$ 0.43 & 91.13 $\pm$ 0.36 & 91.05 $\pm$ 0.46 & 91.17 $\pm$ 0.39 \\
    \multicolumn{2}{l|}{F1} & 86.03 $\pm$ 0.37 & 86.59 $\pm$ 0.33 & 92.01 $\pm$ 0.42 & 91.96 $\pm$ 0.45 & 92.21 $\pm$ 0.33 & 92.09 $\pm$ 0.37 \\
    \multicolumn{2}{l|}{UAR} & 81.14 $\pm$ 0.33 & 83.77 $\pm$ 0.38 & 88.87 $\pm$ 0.49 & 88.91 $\pm$ 0.46 & 88.96 $\pm$ 0.33 & 89.15 $\pm$ 0.35 \\
    \bottomrule
  \end{tabular}
  }
\end{table*}

\begin{figure}[t]
    \centering
    \includegraphics[width=0.85\linewidth]{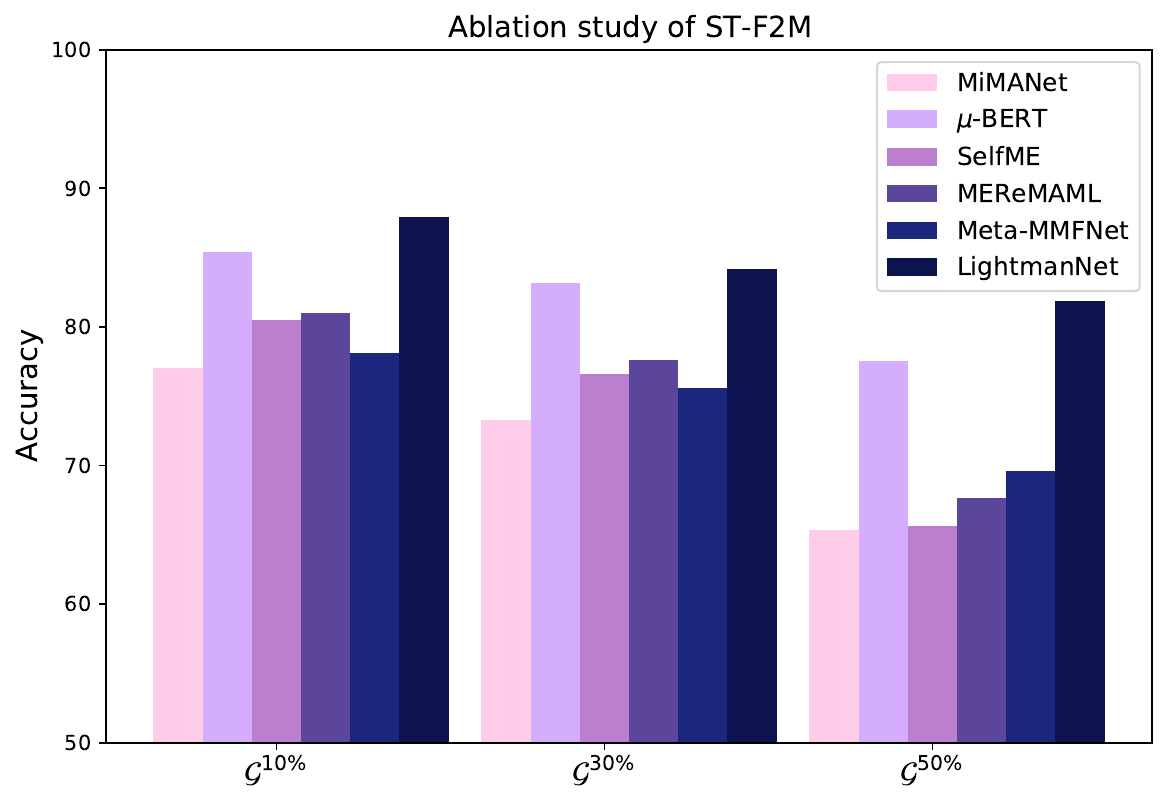}
    \caption{Performance comparison (Accuracy) on the CDE dataset when faces noises. The $\mathcal{G}^{10\%}$, $\mathcal{G}^{30\%}$, and $\mathcal{G}^{50\%}$ respectively represent samples faced with different proportions of noise, e.g., the results of $\mathcal{G}^{10\%}$ are the average results of three experiments on the CDE dataset with $\mathcal{P}_{1}^{10\%} $, $\mathcal{P}_{2} ^{ 10\%} $, and $\mathcal{P}_{3} ^{10\%}$.}
    \label{fig:robust}
\end{figure}

\begin{figure}[t]
    \centering
    \includegraphics[width=0.80\linewidth]{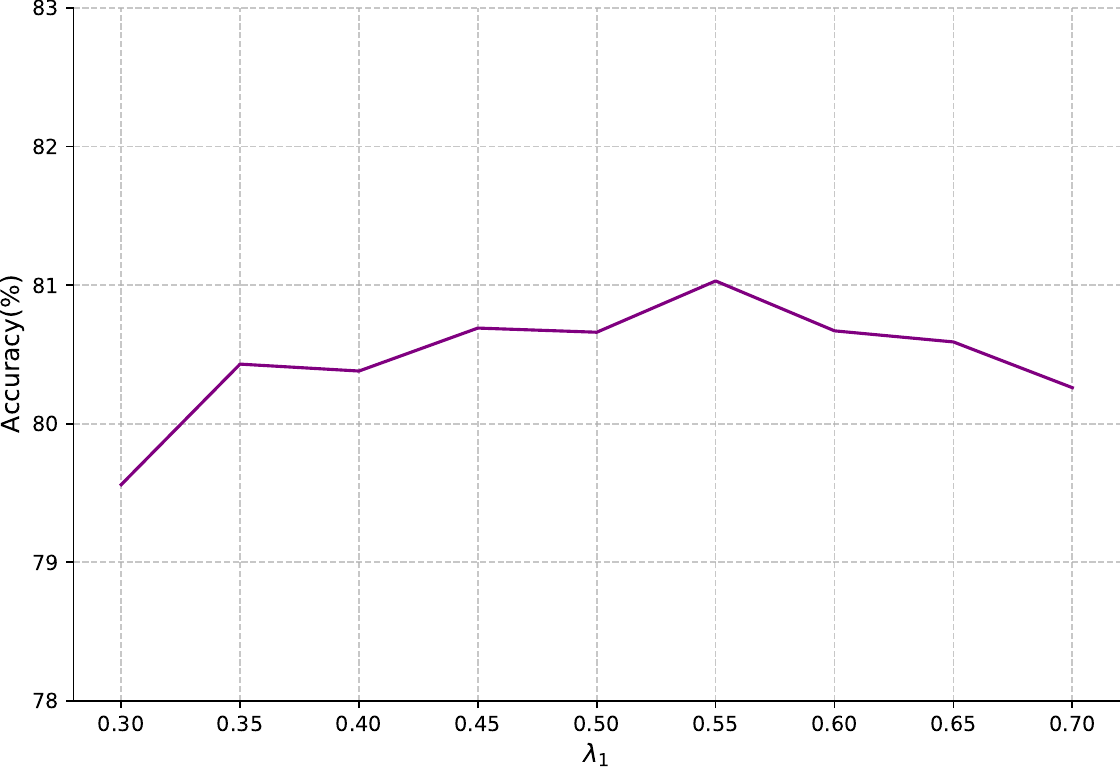}
    \caption{The ablation study of parametric sensitivity, i.e., the effect of hyperparameters $\lambda_1$ and $\lambda_2$ on model performance at different values. $\lambda_1$ and $\lambda_2$ are the hyperparameters of two branches' losses mentioned in Subsection \ref{sec:4}. Note that $\lambda_2=1-\lambda_1$, and the horizontal axis represents $\lambda_1$.}
    \label{fig:abla_para}
\end{figure}

\subsection{Comparative Experiments}
\label{sec:5.2}
In this subsection, we introduce the comparative experiments conducted to evaluate the effectiveness of LightmanNet, including comparisons of performance and model efficiency.

\paragraph{Performance}
We conduct comparison experiments about the performance of LightmanNet and all the baselines on five datasets mentioned in Subsection \ref{sec:5.1}. We use consistent experimental settings for each baseline, including the backbone network, hardware platform, etc. The quantitative results are illustrated in Table \ref{tab:comparison} and Figure \ref{fig:test}.
From Table \ref{tab:comparison}, we observe that LightmanNet achieves state-of-the-art (SOTA) accuracy and F1 score on all the benchmark datasets. For example, LightmanNet reaches an accuracy of 81.83\% in dataset SAMM, surpassing the second-best method by 5.13\%, while achieving an improvement of more than 3\% in both accuracy and F1 scores on CASME dataset. Furthermore, Figure \ref{fig:test} shows the UAR comparison results on the CDE dataset, which reflects the model performance on various tasks and domains. As expected, LightmanNet performs better than almost all the MER baselines. In summary, all the results demonstrate the outstanding effectiveness and robustness of our proposed method.

\paragraph{Model Efficiency}
To ensure the effectiveness of the method in practical applications, we evaluate the model efficiency from three aspects: performance (accuracy \%), training time (training hours, h), and model size (parameter size, M). Figure \ref{fig:model_eff} shows the trade-off performance of various baselines and our LightmanNet. The horizontal axis represents the training hours and the vertical axis represents the accuracy. The center of each circle represents the result of the training time and accuracy of each model, and the area of the circle represents the model size. From the results, we can observe that: (i) LightmanNet achieves performance far exceeding traditional MER methods with acceptable computation time and model size. (ii) Compared with the meta-learning-based and deep-learning-based baselines, LightmanNet achieves better results in a shorter time cost. The results demonstrate the outstanding model efficiency and advantages in real-world applications of LightmanNet.

\subsection{Robust Analysis}
\label{sec:5.3}
Considering that the real-world MER faces problems of data scarcity and imbalanced class (one MER challenge mentioned in Section \ref{sec:1}), we conduct robust analyses to evaluate the model's performance on two challenging MER scenarios, i.e., the few-shot MER and MER with noises where the real-life applications will face. 

\paragraph{Few-shot MER} To create few-shot tasks, we randomly choose $K$ samples (each comprising $M$ frames) from $N$ classes, ensuring no overlap between classes within each task. Here, $K=1$ or $K=5$ shows the meaning of few-shot learning, and each frame contains the label of macro-expressions and micro-expressions. This results in an $N$-way $K$-shot task, featuring $N$ classes (different emotions of different individuals) and $N \times K$ samples. Before training, we construct $N_{tr}$ training tasks through the above process and then train the model via the constructed training tasks. After training, we evaluate the trained model's performance on unseen MER tasks by repeating the above random sampling process.

The results are shown in Table \ref{tab:few-shot}. From the results, we can observe that across almost all the datasets, LightmanNet consistently outperforms both meta-learning methods and MER methods. This indicates that LightmanNet can achieve similar or even superior generalization improvements compared to baselines, without the need for extensive data reliance in the real world.

\paragraph{MER with noises} In addition to data-level challenges, real-life MER also faces recognition difficulties caused by data quality, such as occlusion, low resolution \cite{naji2022emo,al2018survey}, etc. In order to verify the robustness of LightmanNet, we conduct experiments on the CDE dataset where the data are adjusted via three types of processing, i.e., introducing random mask $\mathcal{P}_{1} $, introducing regional Gaussian noise $\mathcal{P}_{2} $, and performing grayscale $\mathcal{P}_{3} $. Then, we adjust the proportion of the adjusted data in the CDE dataset with 10\%, 30\%, and 50\%, corresponding to the superscript of the operation $\mathcal{P}$, e.g., $\mathcal{P}_{1}^{10\%} $ denotes that 10\% of the data with random masks. Next, we record the three sets of average results on different processing, e.g., the first set $\mathcal{G}^{10\%} $ denotes that recording the model performance on the CDE dataset with $\mathcal{P}_{1}^{10\%} $, $\mathcal{P}_{2} ^{ 10\%} $, and $\mathcal{P}_{3} ^{10\%}$.

The results are shown in Figure \ref{fig:robust}. From the results, we can observe that: (i) the performance of LightmanNet does not change much with all three scales of data corruption, proving its advantages in robustness; (ii) in particular, when the noise ratio reaches 50\%, the performance of LightmanNet only drops less than 5\%, significantly exceeding other methods. These findings verify the robustness and effectiveness of LightmanNet and its advantages in practical applications, where data quality corruption due to acquisition difficulties of MER and artificial occlusion always exists.

\subsection{Ablation Study}
\label{sec:5.4}
In this subsection, we conduct ablation studies to further explore how LightmanNet works. We first conduct experiments on the CDE dataset mentioned in Subsection \ref{sec:5.1} to evaluate the effect of different modules in LightmanNet. Next, we evaluate the parametric sensitivity of $\lambda_1$ and $\lambda_2$, which are the hyperparameters of the two learning branches of LightmanNet.

\paragraph{The effect of different modules} We analyze the impact of the three components of LightmanNet, including the feature extractor, the guidance of auxiliary tasks (the effect of introducing auxiliary tasks), and bi-level optimization. We perform evaluation by activating or deactivating/replacing the corresponding modules. Specifically, for the feature extractor, we try four types of extractors, i.e., Conv4, ResNet, VGG-16, and DenseNet, as mentioned in Subsection \ref{sec:4.1}. For the guidance of auxiliary tasks, we directly disabled the learning of auxiliary tasks and the model performs learning only through the primary branch. For the bi-level optimization, we replace our method with the traditional method but with the same encoder and use the common supervised learning manner, i.e., update the network only once per iteration. Table \ref{tab:abla_module} shows the results. From the results, We can observe that: (i) for the feature extractors, their performance is not much different, and CONV4 is the lightest model which we choose as the extractor to improve efficiency. (ii) the dual-branch learning and bi-level optimization can effectively improve the model performance, achieving improvements higher than 5\% and 7\% respectively. This indicates that our design is foresighted.

\paragraph{Parametric Sensitivity} 
The hyperparameters $\lambda _1$ and $\lambda _2$ are the importance weights of two different branches of learning for the first-level optimization objective as mentioned in Eq.\ref{lightman_objective}. We evaluate the performance (accuracy(\%)) of LightmanNet with different $\lambda_1$ and $\lambda_2$, following the same implementation discussed in Subsection \ref{sec:5.2}. Specifically, we vary $\lambda_1$ in the range of $0.3\sim 0.7$ where $\lambda_2=1-\lambda_1$, and record the classification accuracy of Lightman using a Conv4 on SAMM dataset. The results are shown in Figure \ref{fig:abla_para}. We can observe that the model performance is optimal when $\lambda_1=0.55$ and $\lambda_2=0.45$, which are also the hyperparameter settings of LightmanNet.


\section{Conclusion}
\label{sec:6}
In this paper, we explore three key challenges of micro-expression recognition in real-life applications together, i.e., data-level, feature-level, and decision-making-level issues, which have been ignored or limited in previous works. To address all the key challenges and perform well in the real world, we propose a novel dual-branch meta-auxiliary learning method, called LightmanNet, to achieve robust and fast micro-expression recognition. LightmanNet uses dual-branch bi-level optimization to learn general MER knowledge with limited data, which contains two steps: (i) in the first level, the model acquires two-branch knowledge for each task, where one branch handles the main MER task to learn MER features and the other branch performs an auxiliary task of magnifying and aligning micro-expressions and macro-expressions based on their similarity, guiding the model obtain discriminative features; then (ii) in the second level, the model learn the general MER-related knowledge by refining the learned task-specific knowledge from multiple MER tasks, i.e., optimizing the model to make it perform well on various MER tasks. Extensive experiments on various benchmark datasets demonstrate the outstanding robustness and effectiveness of LightmanNet and its obvious advantages in real-world applications.


\bibliographystyle{ACM-Reference-Format}
\bibliography{sample-base}

\end{document}